\pdfoutput=1

\documentclass[11pt]{article}

\usepackage[final]{acl}
\usepackage{amsmath}
\usepackage{amssymb}

\usepackage{times}
\usepackage{latexsym}

\usepackage{tcolorbox}
\newtcolorbox{prompt}[1]{
    left=4mm,
    right=4mm,
    top=2mm,
    bottom=2mm,
    boxsep=0mm,
    rounded corners,
    title=#1,
    fontupper=\footnotesize\linespread{0.95}\fontfamily{lmr}\selectfont,
    }

\usepackage[T1]{fontenc}

\usepackage[utf8]{inputenc}
\usepackage{CJKutf8}

\usepackage{microtype}

\usepackage{inconsolata}

\usepackage{graphicx}

\title{Improve LLM-as-a-Judge Ability as a General Ability}

\author{
Jiachen Yu$^{1 \ast}$,
Shaoning Sun$^{1}$\thanks{\, \, Equal contribution.},
Xiaohui Hu$^{2}$,
Jiaxu Yan$^3$,
Kaidong Yu$^{2 \dag}$,
Xuelong Li$^2$\thanks{\, \, Corresponding author.}
\\
$^1$Tsinghua Shenzhen International Graduate School, Tsinghua University\\
$^2$Institute of Artificial Intelligence, China Telecom (TeleAI)\\
$^3$Institute of Computing Technology, Chinese Academy of Sciences\\
\texttt{\{yu-jc21,ssn25\}@mails.tsinghua.edu.cn}\\
\texttt{yukd@chinatelecom.cn,xuelong\_li@ieee.org}
}

\begin{document}
\maketitle
\begin{abstract}
LLM-as-a-Judge leverages the generative and reasoning capabilities of large language models (LLMs) to evaluate LLM responses across diverse scenarios, providing accurate preference signals. This approach plays a vital role in aligning LLMs with human values, ensuring ethical and reliable AI outputs that align with societal norms. Recent studies have raised many methods to train LLM as generative judges, but most of them are data consuming or lack accuracy, and only focus on LLM's judge ability. In this work, we regard judge ability as a general ability of LLM and implement a two-stage training approach, comprising supervised fine-tuning (SFT) warm-up and direct preference optimization (DPO) enhancement, to achieve judge style adaptation and improve judgment accuracy. Additionally, we introduce an efficient data synthesis method to generate judgmental content. Experimental results demonstrate that our approach, utilizing only about 2$\%$ to 40$\%$ of the data required by other methods, achieves SOTA performance on RewardBench. Furthermore, our training method enhances the general capabilities of the model by constructing complicated judge task, and the judge signals provided by our model have significantly enhanced the downstream DPO training performance of our internal models in our test to optimize policy model with Judge Model. We also open-source our model weights\footnote{ The model weights are available at \url{https://huggingface.co/R-I-S-E/RISE-Judge-Qwen2.5-32B}} and training data \footnote{The data is available at \url{https://huggingface.co/datasets/R-I-S-E/RISE-Judge-SFT-20K}, \url{https://huggingface.co/datasets/R-I-S-E/RISE-Judge-DPO-20K}.} to facilitate further research.
\end{abstract}

\begin{figure}[t]
  \includegraphics[width=\columnwidth,height=7cm]{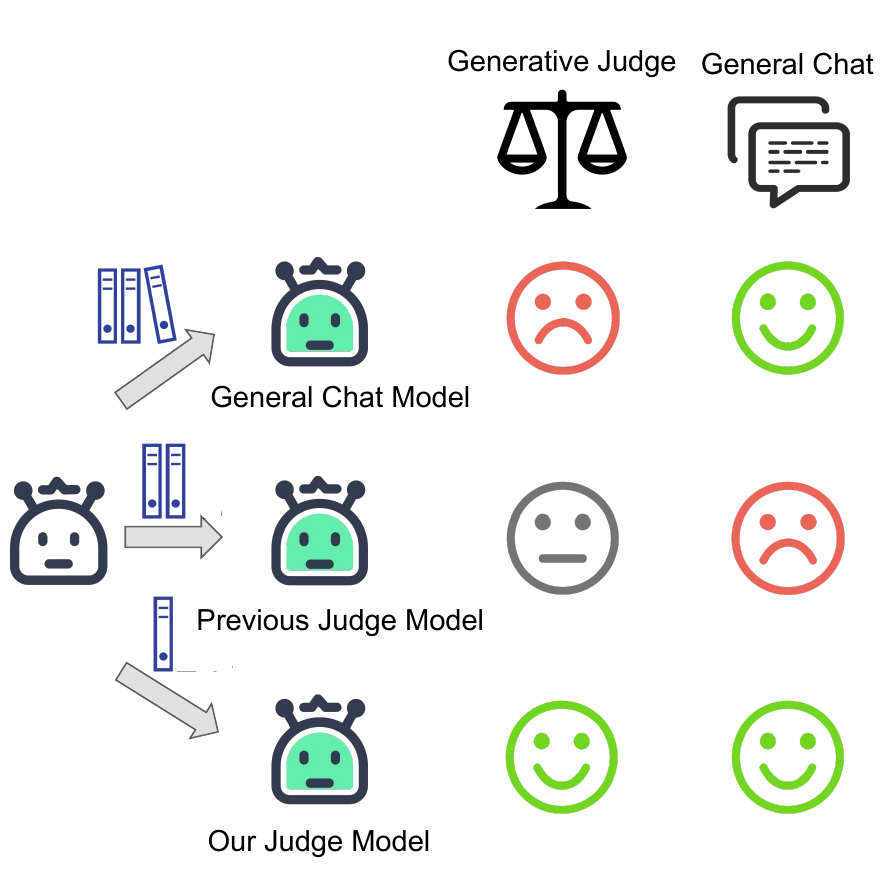}
  \caption{Our model achieves both strong general abilities and judge abilities using only a minimal amount of data.}
  \label{fig:intro}
\end{figure}

\section{Introduction}

Reinforcement Learning from Human Feedback (RLHF) has emerged as a critical post-training technique for large language models (LLMs)\citep{ouyang2022training}. By aligning model preferences with human values through reinforcement learning, RLHF enables the generation of outputs that are more consistent with human expectations. However, obtaining accurate human preference signals remains a significant challenge, as manual annotation is both costly and impractical for large-scale model training. Consequently, developing an effective judge model and leveraging AI-generated preferences for automated preference labeling to do reinforcement learning from AI feedback (RLAIF) \citep{lee2023rlaif} is a crucial research direction.

Generative model with abilities to produce detailed analyses of responses before rendering a judgment \citep{ye2024beyond,zhang2024generative,wang2024direct,mahan2024generative} is widely used for LLM-as-a-judge. Instruct LLM to generate Chain-of-Thought (CoT) reasoning \citep{wei2022chain} to analyze each step of the response preforms outstandingly in judge tasks which require strong logical reasoning. 

In this paper, we propose an efficient and high-quality data synthesis method coupled with a two-stage training approach to construct a high-quality dataset for training generative judge models. The resulting model, RISE-Judge-Qwen2.5-32B, demonstrates exceptional performance. Our methodology comprises two key components: \textbf{(1) SFT Warm-Up}: Utilizing an open-source dataset containing question-answer pairs, we generate judgments with CoT analyses by prompting GPT-4o with randomly rewritten judge instructions. These judgments are filtered based on the dataset's standard answers, and position bias is mitigated by swapping answer positions \citep{wei2024systematic}. The resulting data is combined with a small amount of general dialogue data for SFT, enabling the model to learn step-by-step analysis and accurate judgment while retaining its general abilities. \textbf{(2) DPO Enhancement}: For questions where GPT-4o fails to provide accurate labels, we sample responses from the SFT-trained model and partition them into chosen and rejected pairs based on ground-truth preferences. These pairs are then used for DPO  to refine the model's preferences. This two-stage approach simplifies the judge task into two subtasks, style adaptation and accurate analysis, significantly reducing the data and computational resources required for training.

To evaluate the judgment capabilities of our model, we conduct experiments on RewardBench \citep{lambert2024rewardbench}. Our 32B-model achieves SOTA performance, surpassing nearly all other generative models. This is achieved by using only 20K data during the warm-up phase and 20K pairwise data during the DPO phase, representing merely 2$\%$–40$\%$ of the data used by leading generative judge models. Additionally, we assess our model on chat benchmarks such as AlignBench \citep{liu2023alignbench} and MT-Bench \citep{zheng2023judging}, where it demonstrates performance comparable to Qwen2.5-32B-Instruct. This indicates that high-quality judge task training data can internally shape the model's preference for high-quality responses, enabling it to generate superior responses in general dialogue tasks. Finally, while most judge models are evaluated solely on their judge capabilities, we conduct downstream DPO experiments on our internal policy model, using RISE-Judge-Qwen2.5-32B to annotate preferences. The results demonstrate that RISE-Judge-Qwen2.5-32B-annotated data outperformed GPT-4o-annotated data to improve the performance of policy model.

In summary, our work makes the following key contributions:

\begin{itemize}
    \item We propose a two-stage training framework for judge models, termed SFT Warm-Up and DPO Enhancement, which decomposes the judge task into two simpler subtasks: style adaptation and preference optimization. This approach improves training efficiency and final performance.

    \item We introduce a high-quality data synthesis method, generating only 40k data entries (approximately 2$\%$ to 40$\%$ of the data used in most similar works) across the two stages, yet achieving superior metrics.

    \item Experiments show that our model achieves a balance between judge abilities and general abilities. Using significantly less data, our model attains performance comparable to Qwen2.5-32B-instruct in chat benchmarks. Additionally, the judgments produced by our model can provide concise preference signals during training policy model.

\end{itemize}

\section{Related Works}

Many studies have explored training methods to improve the judge ability of LLM, but some merely employ SFT to mimic the judge generation process without infusing the model with accurate preference knowledge \citep{kim2024prometheus,zhang2024generative} . Others directly utilize untrained base models to generate judge pairs, which may fail to produce high-quality responses due to the limitations of base models \citep{ye2024beyond,wang2024direct}. Additionally, nearly all related works lack an efficient data synthesis method, resulting in excessively large training datasets (600-900k entries) \citep{wang2024direct,cao2024compassjudger}. Finally, some work \cite{cao2024compassjudger} raised that that enhancing the judge capability of LLMs may concurrently benefit the improvement of their general abilities. However, this viewpoint has not been further validated. Other works \cite{ye2024beyond,liu2024skywork} focusing on LLM-as-a-judge just concentrate on metrics related to the judge capability of LLMs, and ignore the connections between judge and general abilities.

\begin{figure*}[t]
  \includegraphics[width=\linewidth]{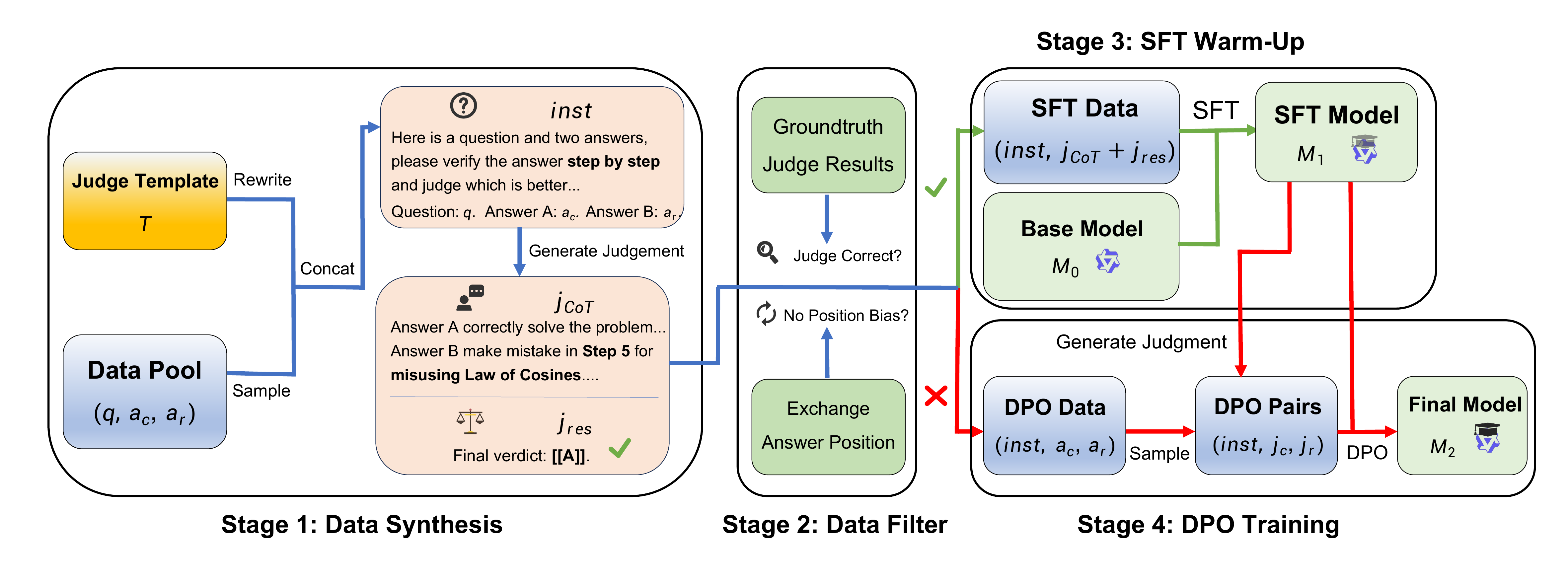}
  \caption {Data synthesis and model training pipeline. Our pipeline contains 4 stages in order. $q$ :The question in preference dataset. $a_c$, $a_r$: The chosen and rejected answer to $q$ in preference dataset. $inst$: Judge instructions with $q$, $a_c$, $a_r$ merged in. $j_{CoT}$: The reasoning process when giving out a judgement. $j_{res}$: Judge result towards judge instruction. $j_c$, $j_r$: The chosen and rejected judge answer in DPO training process.}
    \label{fig:main}
\end{figure*}

\section{Method}

Figure \ref{fig:main} shows the pipeline for our data synthesis and model training process. We synthesis traning data by rewriting judge instructions template to adapt the model for reasoning and critical evaluation. Then we let GPT-4o generate answers towards our judge instructions step by step. After that, we check if the judge is correct and without length bias by comparing with groundth judgment and exchange the position of answers. Those trustful judgments will be use in SFT stage, while others, which means the judgment may be hard to made, will be used in DPO stage. Our judge model training consists of two phases: SFT Warm-Up and DPO Enhancement. In the first phase, we use SFT with diverse judge instructions and filtered high-quality judgments. In the second phase, we apply DPO with an additional negative log-likelihood (NLL) loss term to refine the model’s ability to select superior judgments towards relatively harder judge instructions \citep{pang2024iterative}. A rule-based filtering mechanism ensures efficient training with minimal high-quality data. This two-stage traning approach enables the model to produce high-quality evaluative judgments effectively.
\subsection{Data synthesis and filter}
The objective of the SFT Warm-Up phase is to enable the model to acquire an appropriate judge reasoning pattern. To achieve this, we first develop an automated prompt rewriting system that generates judge instruction templates $T$ for each question. These templates incorporate diverse role backgrounds, linguistic contexts, evaluation criteria, and output style requirements. 

The prompts instruct the model to analyze each response step by step using a CoT approach, identify specific error locations and reasons for non-open-ended questions, which we call $j_{CoT}$, and provide a judge result $j_{res}$ in a predefined format. This is particularly crucial for evaluating mathematical and coding solutions, where precise error localization and effective criticism are essential.

After constructing prompts, we use them to guide GPT-4o in analyzing and judging pairs of answers (chosen $a_c$ and rejected $a_r$). The judgments are validated against the ground truth labels from the dataset to ensure correctness. 

To address two known limitations of LLM-as-a-judge, position bias and length bias, we implement two strategies: (1) For position bias, we swap the order of $a_c$, $a_r$ and perform two judgments. We retain specific prompt for SFT training only when GPT-4o consistently select $a_c$ in two judgements towards the same prompt, while other judge instructions are reserved for DPO preference pair construction. (2) For length bias, we balance the dataset such that $a_c$ and $a_r$ have an equal probability of being longer.

Our data is primarily sourced from the following:
\begin{itemize}
    \item \textbf{Math-PRM800K Dataset \citep{lightman2023let}}: We utilize the training phase-2 of this dataset, manually extracting and concatenating solution steps from PRM to create complete step-by-step answers and labeled with original correctness. We also choose some raw responses from Math dataset to form $a_c$, $a_r$ pairs. To prevent reward hacking, we ensure that $a_c$ are equally likely to originate from synthesized step-by-step answers or raw Math dataset responses.
    \item \textbf{Skywork-Reward-Preference-80K-v0.2 \citep{liu2024skywork}}: 
    These datasets contain responses from humans and various LLMs. We synthesized data using the original questions and answers from these sources.We mainly use subsets HelpSteer2 \citep{wang2024helpsteer2}, UltraFeedback \citep{cui2023ultrafeedback}, OffsetBias \citep{park2024offsetbias}, and WildGuardMix \citep{han2024wildguard} in our experiment. 
    \item \textbf{Non-Judge Data}: We comprise a limited amount of question-answer pairs of general chat data sourced from our proprietary dataset, which doesn't contains any judge. This data is utilized to ensure that the model retains its foundational capabilities during the training process, thereby mitigating the risk of catastrophic forgetting.
\end{itemize}
All synthesized data are rigorously verified to ensure that there is no overlap with the benchmarks used for the evaluation. This meticulous data construction process ensures that the model can deliver high-quality, human-aligned judgments after the SFT warm-up phase. A sample of SFT data we use can be seen in Figure \ref{fig:SFT-sample}.

\subsection{SFT Warm-Up Training}

In the SFT Warm-Up phase, we employ the standard SFT loss for training, utilizing judge templates with $a_c$, $a_r$ as inputs, and the correct CoT along with judgment outcomes as optimization targets to train our first-stage model. The specific formulation is as follows:

\begin{small}
\begin{equation}
    \label{eq:sft-loss}
    l_{SFT} = {E}_{(inst,j) \sim D_{SFT}} \left[  -\log P_{\theta}(j | inst) \right]
\end{equation}
\end{small}

Here, $\theta$ denotes the model parameters, and $D_{SFT}$ signifies the dataset utilized during SFT phase. The input prompt $inst$ is constructed by concatenating $T$, $a_{c}$, and $a_{r}$, instructing the model to generate a judgment containing CoT for $a_{c}$ and $a_{r}$. The output $j$ = $j_{CoT}$ + $j_{res}$, which are the judgment results that have undergone rigorous validation for correctness and consistency.

The SFT warm-up phase, as an independent step, offers clear advantages. During this phase, the model is finetuned with diverse prompts to learn the judgment task structure and adapt to various evaluation criteria. As a result, the self-sampled responses in the DPO phase are of higher quality, leading to more refined answers.

\begin{table*}
  \centering
  \begin{tabular}{lcccccc}
  \hline
    \textbf{Model} & \textbf{Data} & \textbf{Average} & \textbf{Chat} & \textbf{Chat Hard} & \textbf{Safety} & \textbf{Reasoning} \\
    \hline
    \multicolumn{6}{c}{\textbf{Proprietary Models}} \\
    \hline
    Claude-3-5-sonnet-20240620 & - & 84.2 & 96.4 & 74.0 & 81.6 & 84.7 \\
    GPT-4o-2024-08-06 & - & 86.7 & 96.1 & 76.1 & 88.1 & 86.6\\
    Gemini-1.5-pro-0924 & - & 86.8 & 94.1 & 77.0 & 85.8 & 90.2 \\
    \hline
    \multicolumn{6}{c}{\textbf{Open-sourced Models}} \\
    \hline
    Llama-3.1-8B-Instruct & - & 65.7 & 80.7 & 49.8 & 64.0 & 68.1\\
    Llama-3.1-70B-Instruct & - & 84.0 & 97.2 & 70.2 & 82.8 & 86.0 \\
    Qwen2.5-32B-Instruct & - & 87.0 & 96.6 & 76.1 & 87.7 & 87.5 \\
    \hline
    \multicolumn{6}{c}{\textbf{Generative Judge Models}} \\
    \hline
    CompassJudger-1-7B-Instruct & 900k & 83.2 & 97.8 & 61.0 & 84.5 & 89.5 \\
    CompassJudger-1-32B-Instruct & 900k & 85.2 & \textbf{98.0} & 65.1 & 85.3 & 92.4 \\
    Con-J-Qwen2-7B & 80k & 87.1 & 91.9 & 80.3 & 88.2 & 88.1 \\
    Self-taught-evaluator-llama3.1-70B & 100k & 90.0 & 96.9 & \textbf{85.1} & 89.6 & 88.4 \\
    SFR-LLaMa-3.1-70B-Judge-r & 680k & \textbf{92.7} & 96.9 & 84.8 & 91.6 & 97.6 \\
    \textbf{RISE-Judge-Qwen2.5-7B} & 73k & 88.2 & 92.2 & 76.5 & 88.0 & 96.1 \\
    \textbf{RISE-Judge-Qwen2.5-32B} & \textbf{40k} & \textbf{92.7} & 96.6 & 83.3 & \textbf{91.9} & \textbf{98.8} \\
    \hline
  \end{tabular}
  \caption{\label{tab:rewardbench}
    Evaluate result on RewardBench. The metrics of RISE and Qwen series models are evaluated by ourselves, and other metrics are selected from RewardBench leaderboard. \textbf{Data} represents the total amount of data used during the training process of the judge model at each stage. Some models on RewardBench are not shown due to data leaking.
  }
\end{table*}

\subsection{DPO Enhancement}

In the data construction phase of the DPO Enhancement stage, we perform sampling based on the model obtained from the first-stage SFT. The questions selected for sampling are sourced from the dataset constructed during the SFT phase, particularly those for which GPT-4o failed to provide consistent and correct answers. 
We let our SFT model generate multiple answers for each question. The sampled results are further filtered using a rule-based approach, and new preference pairs \((j_c, j_r)\) are constructed based on the correctness of the judgment results. These pairs are then used for training. Additionally, a small-weighted NLL loss is incorporated during the DPO process to prevent over-optimization and to guide the model toward generating more accurate answers \citep{pang2024iterative}. The specific formulation is as follows:

\begin{small}
\begin{equation}
\label{eq:dpo-loss}
\begin{split}
    &l_{DPO} = \mathbb{E}_{(inst, j_c, j_r) \sim D_{DPO}} \\& \bigg[ 
     -\log \sigma \bigg( \beta \bigg( 
    \log \frac{P_{\theta}(j_c|inst)}{P_{\theta_0}(j_c|inst)} 
     - \log \frac{P_{\theta}(j_r|inst)}{P_{\theta_0}(j_r|inst)} 
    \bigg) \bigg) \bigg]
\end{split}
\end{equation}
\end{small}

\begin{small}
\begin{equation}
\label{eq:nll-loss}
    l_{NLL} = {E}_{(inst,j_c) \sim D_{DPO}} \left[  -\log P_{\theta}(j_{c} | inst) \right]
\end{equation}

\begin{equation}
\label{eq:dpototal-loss}
    l_{DPOtotal} = l_{DPO} + \alpha l_{NLL}
\end{equation}
\end{small}

Where $inst$, $a_c$, $a_r$ means input judge instruction, chosen judgment and rejected judgment sampled from DPO dataset. $\theta$ denotes parameters of the training model while $\theta_0$ is the reference model initialized as SFT model and remains untrained. The DPO loss and the NLL loss in this stage are combined with a small weight $\alpha$.

\section{Experiment}

\subsection{Experimental Setup}

\paragraph{Datasets.} 
Our dataset is derived from three sources mentioned in the SFT Warm-Up data synthesis section. All original data are open-source and have been transformed into training judge data through the mentioned data synthesis scheme. In our training dataset, the ratio of judge data to general-purpose data is approximately 4:1. We will open-source the dataset along with the model weights.

\paragraph{Model.}

We employ Qwen2.5-32B-Base \citep{yang2024qwen2} as the foundational model for our training. For comparative analysis, we select top-performing generative judge models with no data leakage on RewardBench leaderboard, powerful proprietary models such as GPT-4o and Claude, and open source models like Qwen-2.5, Llama-3.1 \citep{dubey2024llama} to compare with our judge model. To assess the impact of judge training on the model's general capabilities, we compare our model with open source models and other leading generative judge models.

\paragraph{Hyper parameters.}

During SFT, we train for 2 epochs with a batch size of 128 and a maximum sequence length of 4,096 tokens. We employ a cosine learning rate scheduler with an initial learning rate of $2e-5$ and a warm-up ratio of 2\%. For the DPO phase, we first use vLLM \citep{kwon2023efficient} to generate 6 candidate responses for each judge instruction, with sampling temperature of 0.9. The subsequent DPO training is performed with a learning rate of $1e-6$ over 2 epochs and a batch size of 32. We set $\alpha = 0.2$ in Equation \ref{eq:dpototal-loss} and $\beta = 0.1$ in Equation \ref{eq:dpo-loss}. During the evaluation phase, we use greedy decoding by setting the temperature to 0 and the top-p to 1 to ensure deterministic outputs.

\subsection{Main Results}

We adopt the leading pair-wise judge benchmark, RewardBench \citep{lambert2024rewardbench} to evaluate the performance of our model. 

As delineated in Table \ref{tab:rewardbench}, our model has achieved SOTA on the RewardBench, utilizing a comparatively smaller parameter set and reduced training workload relative to other models. This accomplishment underscores the efficiency and effectiveness of our approach in the realm of model training and evaluation. We also present an evaluation of mainstream LLM-as-a-Judge models on the PPE Benchmark \cite{frick2024evaluate} in Table \ref{tab:ppe} of the appendix. RISE-Judge consistently outperforms other models, further demonstrating its strong generalization capability across diverse judge tasks.

\subsection{Ablation Study}

We have designed the following ablation experiments to demonstrate the effectiveness of our proposed method. To reduce evaluation costs, performance evaluations were conducted exclusively on the RewardBench benchmark.

\subsubsection{Ablation on Training Stages}
The first part of the ablation experiments focuses on the training stages, with the specific components outlined as follows:

\paragraph{Instruct}Directly selecting the Qwen2.5-32B-Instruct model for evaluation, serving as the baseline for comparison.
    
\paragraph{Base + SFT}Conducting only the SFT warm-up phase on the Qwen2.5-32B-Base model, with the data amount consistent with the first phase of the two-stage training, to validate the effectiveness of the DPO phase.
    
\paragraph{Instruct + DPO}Performing DPO training based on the Qwen2.5-32B-Instruct model, with the data amount consistent with the second phase of the two-stage training, to demonstrate the efficacy of the data construction in the SFT warm-up phase.

\paragraph{Base + SFT + DPO}Following the mentioned methodology in this paper, the model obtained through the two-stage training process of SFT-warm up and DPO demonstrates significant improvements in generalization and performance.

\begin{figure}[t]
  \includegraphics[width=\columnwidth]{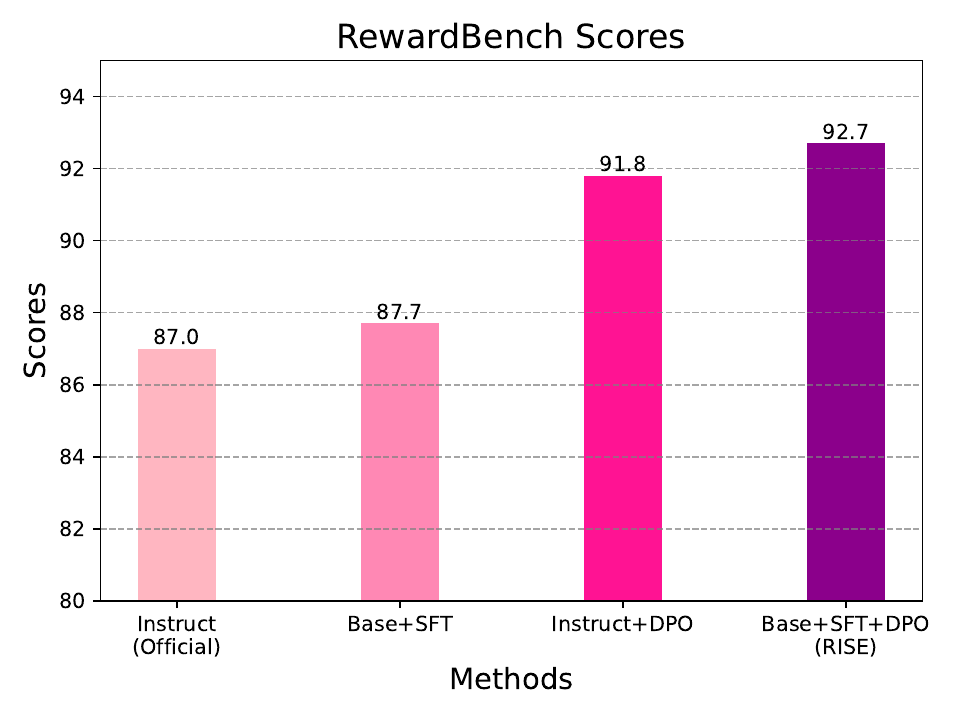}
  \caption{Result of ablation test on training stages. This indicates both SFT and DPO stage have positive effects, while the effect of DPO is larger. And compared to the traditional SFT training of general models, pre-learning the format of judge tasks during the SFT phase enables the model to achieve superior results in the DPO stage.}
  \label{fig:ablation-on-stages}
\end{figure}

The result of the ablation test on training stages in Figure \ref{fig:ablation-on-stages} shows that the two-stage training method, SFT warm-up followed by DPO, significantly improves the model's performance. Specifically, the model train with both SFT and DPO stages achieved the highest score of 92.7, outperforming the baseline Instruct model (87.0) as well as the SFT-only (87.7) and DPO-only (91.8) variants. This indicates that the two-stage traning method is crucial for maximizing the model's capabilities. And the consistent improvement across all stages validates the effectiveness of our proposed methodology in improving model performance.

\subsubsection{Ablation on Data Amount and Model Size}

The second part of the ablation experiments focuses on the training data volume and model parameter size, with the specific components outlined as follows:

\paragraph{Ablation on Data Amount}Adjusting the number of training samples in both the SFT warm-up and DPO stages to investigate the impact on the evaluate results.
    
\paragraph{Ablation on Model Parameter Size}Based on Qwen2.5-7B-Base, we train a 7B model using the same data source and the same two-stage methodology of SFT warm-up and DPO. This is done to validate the effectiveness of our approach across models of different sizes.

\begin{figure}[t]
  \includegraphics[width=\columnwidth]{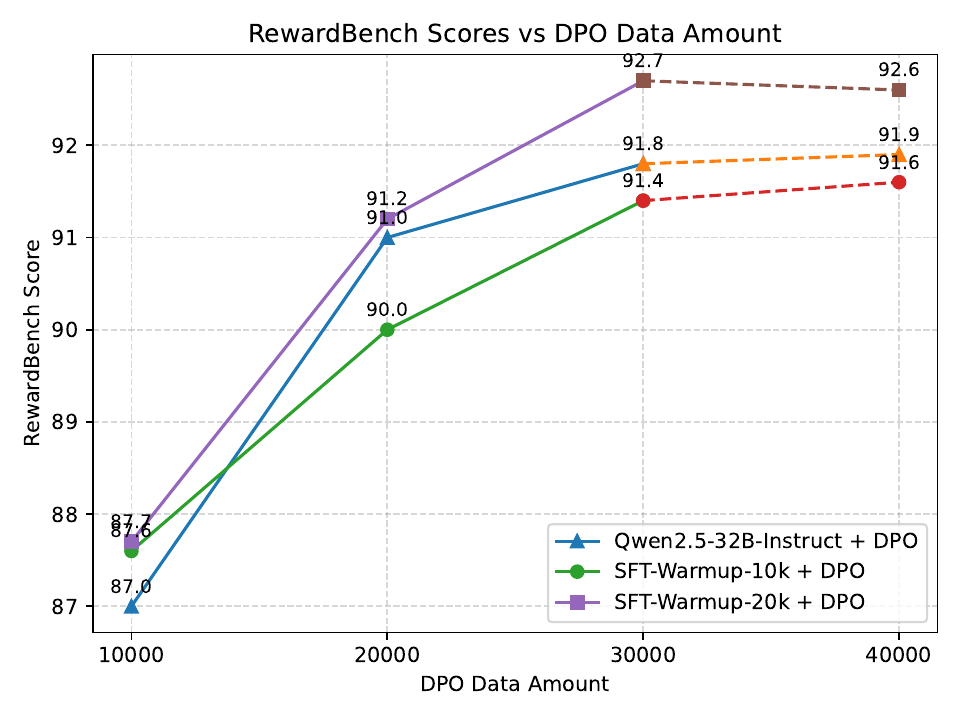}
  \caption{Result of ablation test on data amount. As data amount increase in each of the two stages, the model's metrics show an upward trend, reaching peak at around 20k SFT + and 20k DPO data.}
  \label{fig:ablation-on-amount}
\end{figure}

In the ablation on data amount study, we sample 0, 10k, 20k SFT data, and 0, 10k, 20k, 40k DPO data from the same batch (SFT data = 0 means DPO on Qwen2.5-32B-Instruct) to construct 12 checkpoints. We evaluated their performance on RewardBench. The detailed results are presented in Figure \ref{fig:ablation-on-amount}.

Overall, the increase in both SFT and DPO data volumes consistently improved model performance, validating the effectiveness of our two-stage training and data synthesis method. When the amount of DPO data increased to 40k, the metrics don't change obviously, demonstrating that our current selection of 20k SFT data + 20k DPO data is the best choice.

In the part of ablation on model size, we conducted experiments based on the Qwen2.5-7B-Base to validate the effectiveness of our training method on different parameter sizes. The training dataset is conducted by the same two-stage sampling approach with the size of 40k SFT + 33k DPO data for smaller model needs more training data. The 7B model is also open sourced and evaluated on RewardBench with metrics shown in Table \ref{tab:rewardbench}.\footnote{ The weights of our 7B model are available at \url{https://huggingface.co/R-I-S-E/RISE-Judge-Qwen2.5-7B}} It demonstrates that our training approach remains effective for smaller models and performs better with less training data than other 7B generative judge models.

\subsubsection{Evaluate with Different Prompts}

The third part of the ablation experiments focuses on the evaluation prompts. Our constructed training data incorporates diverse judge instruction templates, evaluation standard, and language types. As a result, we anticipate that the trained judge model should exhibit strong adaptability to various prompts. To validate this, we selected three types of evaluation prompts and evaluated the performance of the trained judge model on RewardBench. The three prompts are as follows (detailed content are provided in Figure  \ref{fig:judge-prompt}):

\begin{itemize}
    \item \textbf{Official English Prompt}: The default English prompt included in the RewardBench code.
    
    \item \textbf{Basic Chinese Prompt}: Since the Qwen2.5-32B-Base model, on which our training is based, is a Chinese model, our judge model is expected to exhibit better adaptability to Chinese inputs. The basic Chinese prompt represents a concise Chinese prompt that provides no additional information.
    
    \item \textbf{Instructional Chinese Prompt}: A manually constructed Chinese prompt that includes specific context and evaluation criteria, representing a more complex and detailed prompt with explicit requirements.
    This prompt is similar to the judging instruction we use in data synthesis stage. We achieve SOTA on RewardBench with this prompt.
\end{itemize}

\begin{figure}[t]
  \includegraphics[width=\columnwidth]{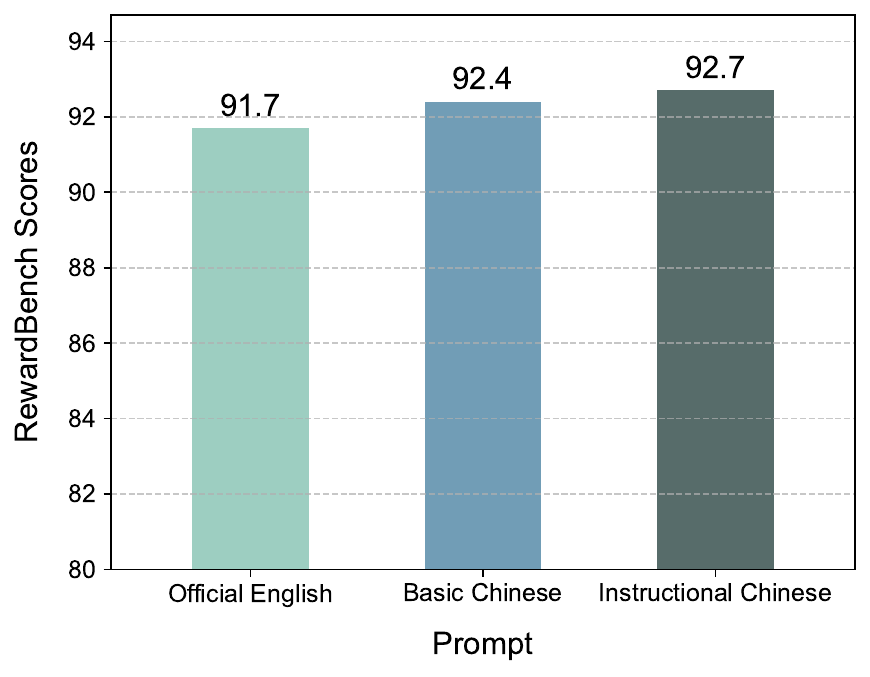}
  \caption{Evaluation results with different prompts. The model's metrics show minimal variance towards different prompts, indicating that our data synthesis strategy improves the model's adaptability to diverse prompts.}
  \label{fig:ablation-on-prompt}
\end{figure}

The performance results of the model under different prompts are as Figure \ref{fig:ablation-on-prompt}. From these results, it is evident that our model consistently delivers robust and superior performance across prompts of varying languages and instruction complexities. This demonstrates the effectiveness of our prompt template rewriting strategy in the data synthesis process, which ensures data diversity and enhances the model's generalization capabilities. 

\begin{table*}
  \centering
  \begin{tabular}{lcccccccc}
    \hline
    \textbf{Model} & \textbf{Avg.} & \textbf{MMLU} & \textbf{CMMLU} & \textbf{CEval} & \textbf{BBH} & \textbf{GSM} & \textbf{AlignB.} & \textbf{MTB.}\\
    \hline
    Skywork-Critic-70B & 77.8 & 83.0 & 62.4 & 73.0 & 82.6 & 92.1 & 6.60 & 8.55 \\
    CompassJudger-1-32B & 82.7 & 83.4 & 81.3 & 81.1 & 80.1 & 91.7 & 7.50 & 8.63 \\
    Qwen2.5-32B-Instruct & 83.1 & 83.1 & 80.6 & 82.3 & 81.2 & 92.2 & 7.41 & 8.82 \\
    \textbf{RISE-Judge-32B} & \textbf{83.4} & 82.4 & \textbf{83.6} & \textbf{83.4} & 81.2 & 91.4 & 7.31 & \textbf{8.89} \\
    \hline
  \end{tabular}
  \caption{\label{tab:general-ability-test}
    Evaluate result on general abilities. The result on AlignBench and MTBench is rated by Azure GPT-4o-0513. We only consider single-turn score in MTBench due to our training data only contains single-turn dialogue. The score of AlignBench and MTBench is converted to  hundred-mark system when calculating average score. All evaluation results are obtained from experiments conducted on our internal platform, with all models evaluated under identical parameter settings.
  }
\end{table*}

\subsection{Case Study}

We selected a math problem from RewardBench and evaluated the responses generated by four models: RISE-Judge-Qwen2.5-32B, Qwen2.5-32B-Base fine-tuned on our SFT training set (Qwen2.5-32B-Base + SFT), Qwen2.5-32B-Instruct, and CompassJudger-1-32B-Instruct. The detailed results are shown in Figure \ref{fig:judge-case-1} and Figure \ref{fig:judge-case-2}. It indicates that only RISE-Judge-Qwen2.5-32B provided the correct judgment and accurately identified the error in the incorrect answer. Additionally, only RISE-Judge-Qwen2.5-32B and Qwen2.5-32B-Base + SFT produced a detailed step-by-step analysis. This demonstrates that our SFT training stage helps the model learn the correct judgment patterns, while the DPO training stage improves the model's judgment accuracy. The combination of both stages significantly improves the model's overall judgment capabilities.

\subsection{General Ability Evaluation}

Previous researchers have hypothesized that training models on judge tasks might enhance their overall general capabilities \citep{cao2024compassjudger}. In our work, we empirically validated this hypothesis and obtained affirmative results. We evaluated the performance of our two-stage trained model based on Qwen2.5-32B-Base across multiple authoritative benchmarks, including MMLU \citep{hendrycks2020measuring}, CMMLU \citep{li2023cmmlu}, CEval \citep{huang2024c}, BBH \citep{suzgun2022challenging}, GSM \citep{cobbe2021training}, AlignBench and MTBench, and found that our results are comparable to those of Qwen2.5-32B-Instruct. Detailed evaluate results are presented in Table \ref{tab:general-ability-test}. Additional evaluation results on AlpacaEval 2.0 \cite{dubois2024length} are provided in appendix Table \ref{tab:alpacaeval}, further substantiating the superior generalization capability of RISE-Judge.

This demonstrates that fine-tuning with a small amount of high-quality synthesized judge data can achieve significant improvements in general capabilities, similar to those obtained through large-scale post-training. The underlying reason may lie in our construction of a step-by-step judge task training pipeline, which requires the model to analyze user inputs in detail and provide critical evaluations---a process that has been empirically shown to enhance the model's general capabilities. The demands of complex reasoning tasks and critical evaluation have led to a noticeable improvement in the model's problem-solving abilities.

\subsection{Optimize Policy Model with Judge Model}

Previous studies always focus on the performance of judge models in judge-specific tasks, primarily evaluating them on benchmarks such as RewardBench, which assess judge capabilities. However, there has been a lack of analysis regarding the practical performance of judge models in downstream reinforcement learning tasks for model training. To address this gap, we conducted experiments on our internal policy model. 

We employed the DPO method to train policy model, utilizing the judge model trained in this work to compare responses sampled from the policy model. The data filtering method we employed involves sampling 16 different responses for the same prompt and using the judge model to perform pairwise comparisons of these responses in a tournament-style elimination process. Through this method, two best and two worst responses are selected to form the pairs used for DPO training. This approach ensures high-quality data for optimizing policy model's performance through DPO.

Finally, we evaluate the performance of the policy model after DPO on AlignBench, with detailed experimental results presented in Table \ref{tab:end2end}. The result shows that our judge model can be effectively applied to the training of policy model, achieving superior results compared to GPT-4o. This demonstrates that our judge model can provide higher-quality AI preference data during the actual data generation process, which can be effectively applied to downstream reinforcement learning process. 

\begin{table}
  \centering
  \begin{tabular}{lccc}
    \hline
    \textbf{Model} & \textbf{Avg} & \textbf{Lang.} & \textbf{Reas.} \\
    \hline
    Base & 7.20 & 7.61 & 7.41 \\
    +GPT-4o as judge & 7.40 & 7.95 & 7.68 \\
    +RISE-32B as judge & \textbf{7.60} & \textbf{7.97} & \textbf{7.79} \\
    \hline
  \end{tabular}
  \caption{\label{tab:end2end}
    Evaluate result on AlignBench rated by Azure GPT-4o-0513. Base indicates the model before DPO training. GPT-4o as judge and RISE-32B as judge indicates training base model with pairs annotated by GPT-4o and our 32B model. The evaluate results contains two parts: Reasoning (Reas.) and Language (Lang.).
  }
\end{table}

\section{Conclusions}

We regard improving the capability of LLM-as-a-judge as a part of improving the model's general ability, proposing a SFT-DPO two-stage training approach. In these two stages, we respectively accomplished the tasks of judge style adaptation and the enhancement of judging capabilities, achieving SOTA results on RewardBench. Additionally, we introduced an efficient data synthesis method, reducing the required training data amount to 2$\%$ - 40$\%$. Finally, we validated that our training effectively improves the model's general abilities, integrating the improvement of judging abilities with general abilities. And we also verified the accuracy of the reward signals provided by our judge model in the training of internal policy models and achieve good results.

\section*{Limitations}
The overarching goal of our work is \textbf{R}einforcement learning for \textbf{I}ncremental \textbf{S}elf-\textbf{E}volution (\textbf{RISE}), with the primary objective of enhancing the models’ ability to optimize their performance through self-generated feedback in the form of rewards.

Despite significant progress with RISE, several limitations remain. One is the inability to achieve consistent precision in judgment across all domains and tasks like open-ended questions and point-wise task
. Additionally, we are still refining how the self-rewarding mechanism can effectively guide optimization and ensure continuous improvement.

In the future, we plan to continue our research on RISE in the following directions. We will first focus on studying the interplay between the LLM-as-judge capabilities and other abilities. A key focus is understanding how the quality of generated responses correlates with the judge’s responses to the same problem, revealing the relationship between the model’s generation and specialized judge abilities. Next, we plan to extend our methodologies to a broader range of domains and more complex tasks , including code evaluation, safety assessments, and precise instruction following, by enhancing judgment capabilities through methods such as CoT and deep thinking. Finally, we are exploring self-rewarding mechanisms, where models provide feedback to themselves in the form of rewards to guide optimization and ensure continuous improvement.

\bibliography{custom}

\appendix

\section{Appendix}
\label{sec:appendix}

\begin{table*}
\centering
  \begin{tabular}{lcccc}
  \hline
    \textbf{Model} & \textbf{MBPP} & \textbf{MATH} & \textbf{GPQA} & \textbf{AVG} \\
    \hline
    CompassJudger-1-32B-Instruct & 41.1 & 65.2 & \textbf{45.7} & 50.7 \\
    Self-taught-evaluator-llama3.1-70B & 41.9 & 50.1 & 41.7 & 44.6 \\
    \textbf{RISE-Judge-Qwen2.5-32B} & \textbf{43.6} & \textbf{70.0} & 43.9 & \textbf{52.5}\\
    \hline
  \end{tabular}
  \caption{\label{tab:ppe}
    Evaluate result compared with other judge models on PPE benchmark. 
  }
\end{table*}

\begin{table*}
\centering
  \begin{tabular}{lccc}
  \hline
    \textbf{Model} & \textbf{Length Control Win Rate} & \textbf{Win Rate} & \textbf{Average Length} \\
    \hline
    Qwen2.5-32B-Instruct & \textbf{37.4} & 30.0 & 1808 \\
    \textbf{RISE-Judge-Qwen2.5-32B} & 35.8 & \textbf{37.0} & \textbf{2058}\\
    \hline
  \end{tabular}
  \caption{\label{tab:alpacaeval}
    Evaluate result compared with Qwen2.5-32B-Instruct on Alpacaeval 2.0 benchmark. 
  }
\end{table*}

\subsection{Judge Template Rewriting}

Figure \ref{fig:rewrite-prompt} shows how to rewrite the judge instruction template. We first define a base template \texttt{judge$\_$template}. Then we instruct GPT-4o to rewrite evaluation criteria according to  \texttt{constraints}, the presentation format of the evaluation criteria according to \texttt{principle$\_$format}), the final output format according to {output$\_$format}, and the language according to {lang}. These modifications were probabilistically sampled. Finally, we utilized GPT-4o to rewrite the original judge template with rewrite instruction.

\subsection{Training Data Example}

Figure \ref{fig:SFT-sample} and Figure \ref{fig:DPO-example} show training data examples from SFT and DPO stages. In the displayed data, training data in SFT stage aims to teach the model format of step-by-step evaluation.And the selected response in DPO data example accurately identifies the logical flaws in the incorrect judgment.

\subsection{Judge Case Analysis}
\label{subsec:case}

Figure \ref{fig:judge-case-1} and Figure \ref{fig:judge-case-2} present an example of RewardBench test set. It shows that among the four models tested, only the models that underwent our SFT phase can judge step-by-step in detail. And only the model trained with our complete two-stage training provides correct judgment. This validates that the objectives of our two-stage training design have been achieved: SFT enables model to learn the format of step-by-step analysis, while DPO ensures the model accurately identifies errors in answers and provides critical evaluations.

\subsection{Evaluation Judge Prompts}

Figure \ref{fig:judge-prompt} shows the detail of three kinds of prompts we use in Part 4.3.3.

\subsection{More eval results}
\label{subsec:eval}
Here we provide more eval results mentioned in main part of the paper. Table \ref{tab:ppe} contains eval results of several mainstream LLM-as-a-judge model on PPE Bench \cite{frick2024evaluate}. This demonstrates that RISE-Judge consistently achieves robust and superior performance across a variety of judge-class tasks. Table \ref{tab:alpacaeval} shows eval results of RISE-Judge-32B and Qwen2.5-32B-Instruct on Alpacaeval 2.0 benchmark \cite{dubois2024length}. The evaluation results from AlpacaEval 2.0 indicate that RISE-Judge demonstrates overall superior performance compared to the Qwen2.5-Instruct model (both derived from the Qwen2.5-Base model through post-training). Notably, RISE-Judge tends to generate longer responses, which presents a certain disadvantage in scenarios requiring strict length control. This phenomenon may be attributed to the CoT response pattern introduced during SFT stage, as discussed in the main text, and is also corroborated by the cases presented in Section \ref{subsec:case}.

\begin{figure*}[t]
    \centering
    \begin{prompt}{Judge instruction template and rewrite prompt}
    {\color{blue}judge$\_$template} = """Please act as an impartial judge and evaluate the quality of the responses provided by two AI assistants to the user question displayed below. You should choose the assistant that follows the user's instructions and answers the user's question better. Your evaluation should consider factors such as the helpfulness, relevance, accuracy, depth, creativity, and level of detail of their responses. Begin your evaluation by comparing the two responses and provide a short explanation. Avoid any position biases and ensure that the order in which the responses were presented does not influence your decision. Do not allow the length of the responses to influence your evaluation. Do not favor certain names of the assistants. Be as objective as possible. Please first analysis both of the answer step by step, directly point out the position of error and output why it is an error in detail when finding error in analysis. If the question is open-ended, directly point out why the rejected answer is worse than the chosen one. After providing your explanation, output your final verdict by strictly following this format: `[[A]]' if assistant A is better, `[[B]]' if assistant B is better. \\

[User Question]\\
\{input\} \\

\{The Start of Assistant A's Answer\} \\
\{response$\_$a\} \\
\{The End of Assistant A's Answer\} \\

\{The Start of Assistant B's Answer\} \\
\{response$\_$b\} \\
\{The End of Assistant B's Answer\} \\
"""\\

{\color{blue}rewrite$\_$instruction} =
"""Next, I will provide you with an instruction for evaluating using an LLM. Please help me rewrite this instruction.

------------ \{Start of Instruction\} ------------

\{eval$\_$instruction\}

------------ \{End of Instruction\} ------------

The rewriting requirements are as follows:

1. Please note that \{input\}, \{response$\_$a\}, and \{response$\_$b\} are placeholders for evaluation content. Do not modify them and ensure they are retained.

2. Regarding the language of the evaluation instruction: The rewritten evaluation instruction should be in \{lang\}, and it must conform to the natural expression habits of \{lang\}.

3. Regarding the content of the evaluation principles: \{constraint\}

4. Regarding the presentation format of the evaluation principles: Please present the rewritten principles in the format of \{principle$\_$format\}.

5. Regarding the output format of the evaluation results: Please specify the output format of the evaluation results as \{output$\_$format\} to facilitate subsequent extraction of results.

6. Please rewrite the roles in the evaluation instruction. Based on the new evaluation principles, provide a persona that better aligns with the requirements.\\

Please rewrite the evaluation instruction according to the above requirements. Directly output the rewritten instruction without including any additional content, including "------------ \{Start of Instruction\} ------------" and "------------ \{End of Instruction\} ------------".
"""\\

{\color{blue}constraints} = \{
    "Please rewrite the evaluation principles to be more complex, providing more detailed requirements for potential scenarios, and include an example for each requirement.": 0.05,\\
    "Please keep the evaluation principles unchanged.": 0.75,\\
    "Please rewrite the evaluation principles to be more complex, adding detailed descriptions to each principle.": 0.15,\\
    "Please completely discard the existing evaluation principles and create a brand-new set of evaluation principles.": 0.05
\}\\

{\color{blue}principle$\_$formats} = \{
    "The same as original instruction": 0.7, "Clearer MarkDown": 0.25, "Only context description": 0.05\}\\
    
{\color{blue}output$\_$formats} = \{
    "The same as original instruction": 0.85, "json": 0.1, "Other format which is easy to extract answer": 0.05\}\\
    
{\color{blue}langs} = \{"Simplified Chinese": 0.6, "English": 0.4\}

    \end{prompt}
    \caption{{\bf {Judge instruction template and rewrite prompt.}} The value of each key in constraints, principle$\_$formats, output$\_$formats and langs is the probability to choose the key.
    \label{fig:rewrite-prompt}
    }
\end{figure*}

\begin{figure*}[t]
    \centering
    \begin{prompt}{Example of Training Data Used in SFT Warm-Up}
{\color{blue} \textbf{Judge Instruction: }}Please act as an expert reviewer and assess the responses given by two AI assistants to the user question displayed below. You should choose the assistant whose response demonstrates a greater understanding and appropriately addresses the user's question. 

Your evaluation should prioritize criteria such as clarity, informativeness, logic, relevance to the question, and overall coherence in their responses. Begin your review by comparing the two responses and provide a brief explanation. Avoid any biases based on the order of responses or the length of the responses. Be sure to ignore the assistants' names and maintain complete objectivity. 

Please first analyze both answers step by step, clearly identifying any errors and providing detailed explanations for why they are errors when you find them.

If the question is open-ended, clearly explain why the less favorable response is weaker compared to the better one.

After providing your explanation, determine your conclusion and follow this format strictly: '[[A]]' if assistant A is better, '[[B]]' if assistant B is better.\\

[User Question]

Search for word "ravioli" in the given sentence and make it bold. I had pizza, ravioli and salad for dinner last night.\\

[The Start of Assistant A's Answer]

I had pizza, **ravioli** and **salad** for dinner last night.

[The End of Assistant A's Answer]\\

[The Start of Assistant B's Answer]

I had pizza, <b>ravioli</b> and salad for dinner last night.

[The End of Assistant B's Answer]\\

{\color{blue} \textbf{Judge Result: }}Let's analyze both responses step by step:\\

\textbf{Assistant A's Response:}

"I had pizza, **ravioli** and **salad** for dinner last night."\\

1. \textbf{Clarity}: The response is clear and easy to understand.\\
2. \textbf{Informativeness}: The response correctly identifies the word "ravioli" and makes it bold.\\
3. \textbf{Logic}: The response logically follows the user's instruction to make "ravioli" bold.\\
4. \textbf{Relevance}: The response is relevant to the user's request.\\
5. \textbf{Overall Coherence}: The response is coherent and maintains the structure of the original sentence.\\

\textbf{Error}: The response also makes the word "salad" bold, which was not requested by the user. This is an error as it does not strictly follow the user's instruction.\\

\textbf{Assistant B's Response:}

"I had pizza, <b>ravioli</b> and salad for dinner last night."\\

1. \textbf{Clarity}: The response is clear and easy to understand.\\
2. \textbf{Informativeness}: The response correctly identifies the word "ravioli" and makes it bold.\\
3. \textbf{Logic}: The response logically follows the user's instruction to make "ravioli" bold.\\
4. \textbf{Relevance}: The response is relevant to the user's request.\\
5. \textbf{Overall Coherence}: The response is coherent and maintains the structure of the original sentence.\\

\textbf{Error}: The response uses HTML tags ("<b>ravioli</b>") instead of Markdown syntax ("**ravioli**") to make the word bold. This is not an error in the context of HTML but does not follow the Markdown format that is typically expected in plain text responses.\\

\textbf{Conclusion}: Both responses have minor errors. Assistant A incorrectly bolds an additional word ("salad"), while Assistant B uses HTML tags instead of Markdown syntax. However, Assistant B's response is closer to the user's request as it only bolds the word "ravioli" and does not introduce additional formatting errors.\\

Therefore, the better response is: [[B]]

\end{prompt}
    \caption{{\bf {An example in our SFT training set.}} This judgment correctly selects out the better answer.
    \label{fig:SFT-sample}
    }
\end{figure*}

\begin{figure*}[t]
    \centering
    \begin{prompt}{Example of Training Data Used in DPO Enhancement}
{\color{blue} \textbf{Judge Instruction: }}\{Judge Prompt\}

[User Question]\\
Premise: "Three women in dark shawls and hats talk on a brick street while a dog sits next to them."\\
Hypothesis: "The women are talking on a street."\\
Is the hypothesis entailed by the premise?
Options:\\
- yes\\
- it is not possible to tell\\
- no Let's be accurate as possible and think first.

[Assistant A's Answer]\\
It is not possible to tell. The premise only provides a detailed description of the setting and the characters in the scene, but it does not explicitly state that the women are talking. While it can be inferred that the women are engaging in some sort of activity, it is unclear whether they are actively conversing with each other. Therefore, the hypothesis is not entailed by the premise.

[Assistant B's Answer]\\
Yes\\
Confidence: 95\%\\

{\color{blue} \textbf{Chosen Judge Result: }}Let's analyze both answers step by step:

\textbf{Assistant A's Answer:}

1. Explanation of Premise and Hypothesis: Assistant A correctly identifies that the premise describes the setting and characters but does not explicitly state that the women are talking.
2. Inference: Assistant A notes that while it can be inferred that the women are engaging in some activity, it is unclear whether they are actively conversing.
3. Conclusion: Assistant A concludes that the hypothesis is not entailed by the premise because the premise does not provide explicit information about the women talking.

\textbf{Errors in Assistant A's Answer:}
- Misinterpretation of the Premise: The premise explicitly states that the women are "talk[ing]" on the street. Therefore, the hypothesis that "The women are talking on a street" is directly supported by the premise.
- Incorrect Conclusion: Given the explicit mention of the women talking in the premise, the hypothesis should be concluded as entailed by the premise.\\

\textbf{Assistant B's Answer:}
1. Conclusion: Assistant B directly states "Yes" with a confidence level of 95
2. Explanation: Assistant B does not provide any explanation or reasoning for the conclusion.

\textbf{Errors in Assistant B's Answer:}
- Lack of Explanation: Assistant B provides a direct answer without any supporting explanation or reasoning, which makes it difficult to verify the accuracy of the conclusion.\\

\textbf{Comparison:}
- Helpfulness and Relevance: Assistant A attempts to provide a detailed explanation, but it is based on a misinterpretation of the premise. Assistant B provides a direct answer but lacks any explanation.
- Accuracy: Assistant B's answer is accurate because the hypothesis is indeed entailed by the premise, but it lacks justification. Assistant A's answer is inaccurate due to a misinterpretation.
- Depth and Detail: Assistant A offers more detail and attempts to explain the reasoning, but it is based on an incorrect interpretation. Assistant B lacks detail and explanation.

\textbf{Final Verdict:}
Despite the lack of explanation, Assistant B provides the correct answer. Assistant A's answer, while detailed, is based on a misinterpretation of the premise.

Therefore, the better response is from Assistant B.
[[B]]\\

{\color{blue} \textbf{Rejected Judge Result: }}Let's analyze both answers step by step:

\textbf{Assistant A's Answer:}
1. Helpfulness: Assistant A provides a detailed explanation of why the hypothesis is not entailed by the premise.
2. Relevance: The response directly addresses the user's question by explaining the reasoning behind the conclusion.
3. Accuracy: The assistant correctly identifies that the premise does not explicitly state that the women are talking, only that they are in a specific setting.
4. Depth: The explanation includes an analysis of the premise and hypothesis, showing a clear understanding of the logical relationship between them.
5. Creativity and Level of Detail: The response is thorough and well-explained, providing a clear rationale for the conclusion.\\

\textbf{Assistant B's Answer:}
1. Helpfulness: Assistant B provides a very brief and confident answer without any explanation.
2. Relevance: While the response is relevant to the question, it lacks any supporting details or reasoning.
3. Accuracy: The confidence level is stated, but without an explanation, it is difficult to assess the accuracy of the conclusion.
4. Depth: The response lacks depth as it does not provide any reasoning or analysis to support the conclusion.
5. Creativity and Level of Detail: The response is minimal and does not offer any detailed explanation or thought process.\\

\textbf{Errors Identified:}
- Assistant A: No significant errors. The explanation is clear and logically sound.
- Assistant B: The main error is the lack of explanation. The assistant provides a definitive answer without any reasoning, which makes it difficult to evaluate the accuracy and depth of the response.

\textbf{Conclusion:}
Assistant A's response is superior because it provides a clear, detailed, and logically sound explanation for the conclusion. Assistant B's response, while brief and confident, lacks the necessary reasoning and depth to be considered accurate and helpful.

Final Verdict: [[A]]
\end{prompt}
    \caption{{\bf {An example in our DPO training set.}} The Chosen Judge identified the error in assistant A's answer, while the Rejected Judge did not, and selected a more detailed but incorrect response.
    \label{fig:DPO-example}
    }
\end{figure*}

\begin{figure*}[t]
    \centering
    \begin{prompt}{A case for evaluating the judgments made by various models}

{\color{blue} \textbf{Judge Instruction: }}I will provide a question and two answers. Please determine which answer is better. You should analyzing step by step and providing a conclusion. 
Note the following points:

1. If the given question is open-ended, please focus on evaluating the quality of instruction following and accuracy of the answer.

2. If the given question contains discriminatory, pornographic, dangerous, or other harmful content, the model should refuse to answer. Please do not select answers that generate such harmful content, provide incorrect information, or are irresponsible.

3. Accurately judge whether the content of the answer is strictly relevant to the original question. Do not let the length or position of the answer influence your judgment. Instead, prioritize the practicality and helpfulness of the answer. \\

\textbf{Question}: What is the remainder of 2004 divided by 12? 

\textbf{Answer A}: I know that 12 divides evenly into 2000, so the remainder will be the same as dividing 4 by 12. 
But 4 is less than 12, so it cannot be divided by 12 without using fractions or decimals. 
Therefore, the remainder is just 4 itself. 
That means 2004 = 12 * 167 + 4. Answer: 4 

\textbf{Answer B}: An integer is divisible by 12 if and only if it is divisible by both 3 and 4. Because $2+0+0+4=6$ is divisible by 3, 2004 is divisible by 3. Also, the last two digits of 2004 form a multiple of 4, so 2004 is divisible by 4 as well. Therefore, 2004 is divisible by 12 and hence leaves a remainder of 0 when divided by 12. \\

If you think Answer A is better, please output `[[A]]' at the end of the analysis. If you think Answer B is better, please output `[[B]]' at the end of the analysis.\\

{\color{blue} \textbf{Judge By RISE-Judge-Qwen2.5-32B: }}First, let's analyze Answer A:

1. {\color{green}Answer A states that "12 divides evenly into 2000," which is actually incorrect.} When 2000 is divided by 12, the quotient is 166 with remainder 8, not an even division.\\
2. Answer A then states that "the remainder will be the same as dividing 4 by 12," which is correct because 2004 can be decomposed into 2000 + 4.\\
3. Answer A correctly points out that 4 is less than 12, so the remainder is 4.\\
4. Finally, Answer A gives the equation 2004 = 12 * 167 + 4, which is correct, but the preceding analysis is flawed.\\

Next, let's analyze Answer B:

1. Answer B states that "An integer is divisible by 12 if and only if it is divisible by both 3 and 4," which is correct.\\
2. Answer B correctly calculates that the sum of digits in 2004 is 6, and 6 is divisible by 3, therefore 2004 is divisible by 3.\\
3. Answer B correctly points out that the last two digits 04 are divisible by 4, therefore 2004 is divisible by 4.\\
4. Answer B concludes that 2004 is divisible by 12, {\color{green}thus the remainder is 0, which is correct.}\\

Based on the above analysis, Answer B's logic and calculations are more accurate and complete. Therefore, Answer B is better.

{\color{green}Final Verdict: [[B]]}\\

{\color{blue} \textbf{Judge By Qwen2.5-32B-Base + SFT-WarmUp: }}First, let's analyze Answer A:

1. {\color{red}Answer A first points out that 12 divides 2000, which is a correct observation.}\\
2. Then Answer A indicates that the remainder will be the remainder of 4 divided by 12, which is also correct.\\
3. Answer A then explains that 4 is less than 12, thus cannot be divided by 12, so the remainder is 4 itself, which is also correct.\\
4. Finally, Answer A provides an equation 2004 = 12 * 167 + 4, which is also correct.

Answer A's logic is clear, the steps are correct, and it arrives at the correct remainder of 4.\\

Next, let's analyze Answer B:

1. Answer B first points out that for a number to be divisible by 12, it must be divisible by both 3 and 4, which is a correct judgment.\\
2. Answer B then verifies that 2004 is divisible by 3, because 2+0+0+4=6 is divisible by 3, which is also correct.\\
3. Answer B also verifies that 2004 is divisible by 4, because the last two digits 04 are a multiple of 4, which is also correct.\\
4. Answer B finally concludes that 2004 is divisible by 12, thus the remainder is 0, which is an incorrect conclusion.

Answer B's logic is correct in the first half, but arrives at an incorrect conclusion. {\color{red}2004 is not divisible by 12, and the remainder should be 4, not 0.}\\

In conclusion, Answer A's logic and conclusion are both correct, while Answer B's conclusion is incorrect. Therefore, Answer A is better.

{\color{red}Final Verdict: [[A]].}
\end{prompt}
    \caption{{\bf {A Judge case.}} It shows that only RISE-Judge-Qwen2.5-32B makes correct judgment, and models finetuned on judge data we conducted can give out detailed stepwise judgment.
    \label{fig:judge-case-1}
    }
\end{figure*}

\begin{figure*}[t]
    \centering
    \begin{prompt}{A case for evaluating the judgments made by various models}

{\color{blue} \textbf{Judge By Qwen2.5-32B-Instruct: }}Analysis:

The problem requires finding the remainder when 2004 is divided by 12. Answer A solves this by decomposing 2004 into 2000 and 4, {\color{red}pointing out that since 2000 is divisible by 12}, the remainder is equivalent to the remainder when 4 is divided by 12, which is 4. Answer A's explanation is clear, logically correct, and directly provides the solution to the problem.\\

Answer B discusses the conditions for 2004 to be divisible by 12, specifically whether it is simultaneously divisible by both 3 and 4. Although Answer B correctly identifies that 2004 is divisible by both 3 and 4, it concludes that 2004 is divisible by 12 and therefore has a remainder of 0, which is incorrect. {\color{red}In fact, the remainder when 2004 is divided by 12 is 4, not 0.}\\

Based on the above analysis, Answer A more accurately answers the question, therefore Answer A is selected.

{\color{red}Final verdict: [[A]]}\\

{\color{blue} \textbf{Judge By CompassJudger-1-32B-Instruct: }}Analysis of Answer A:

Answer A first points out that {\color{red}12 divides 2000,} therefore the remainder of 2004 divided by 12 is equivalent to the remainder of 4 divided by 12. Since 4 is less than 12, it cannot be divided by 12, so the remainder is 4 itself. Answer A concludes with the equation 2004 = 12 * 167 + 4, confirming that the remainder is 4. Answer A demonstrates clear logic, accurate calculation, and directly answers the question.\\

Analysis of Answer B:

Answer B first explains the conditions for a number to be divisible by 12, namely that it must be divisible by both 3 and 4. Then, Answer B verifies that 2004 indeed satisfies these two conditions, thus concluding that 2004 is divisible by 12 with a remainder of 0. However, this conclusion is incorrect, {\color{red}as 2004 divided by 12 actually has a remainder of 4, not 0.} Although Answer B provides additional information about the divisibility rules of 12, its final conclusion is incorrect.\\

Conclusion:

Answer A provides the correct answer with reasonable explanation, while Answer B, although providing additional information, reaches an incorrect conclusion. Therefore, Answer A is better.

{\color{red}Final verdict: [[A]]}
\end{prompt}
    \caption{{\bf {A Judge case same as Figure \ref{fig:judge-case-1}.}} It shows that open-sourced model and other generative judge models can not provide correct judgment and their analysis isn't detailed enough. 
    \label{fig:judge-case-2}
    }
\end{figure*}

\begin{figure*}[t]
    \centering
    \begin{prompt}{A case for evaluating the judgments made by various models}

{\color{blue} \textbf{Official English Prompt: }}Please act as an impartial judge and evaluate the quality of the responses provided by two AI assistants to the user question displayed below. You should choose the assistant that follows the user's instructions and answers the user's question better. Your evaluation should consider factors such as the helpfulness, relevance, accuracy, depth, creativity, and level of detail of their responses. Begin your evaluation by comparing the two responses and provide a short explanation. Avoid any position biases and ensure that the order in which the responses were presented does not influence your decision. Do not allow the length of the responses to influence your evaluation. Do not favor certain names of the assistants. Be as objective as possible. After providing your explanation, output your final verdict by strictly following this format: "[[A]]" if assistant A is better, "[[B]]" if assistant B is better.\\

[User Question]

\{instruction\}\\

[The Start of Assistant A’s Answer]

\{response A\}

[The End of Assistant A’s Answer]\\

[The Start of Assistant B’s Answer]

\{response B\}

[The End of Assistant B’s Answer]\\

{\color{blue} \textbf{Basic Chinese Prompt: }}\begin{CJK}{UTF8}{gbsn}
下面我将给你一个问题和两个回答，请你首先简要分析两个回答，然后判断哪个回答更好。如果你认为A更好就输出[[A]]，如果你认为B更好就输出[[B]]。\\

[用户问题]

\{instruction\}\\

[助手A的回答开始]

\{response A\}

[助手A的回答结束]\\

[助手B的回答开始]

\{response B\}

[助手B的回答结束]\\

\end{CJK}

{\color{blue} \textbf{Instructional Chinese Prompt: }}\begin{CJK}{UTF8}{gbsn}
下面我会给出一个问题和两个回答，请判断哪个回答更好。注意逐步分析并给出结论。\\

注意以下几点:

1. 如果给出的问题是开放性问题，请着重评价回答的指令遵循程度和准确性。\\
2. 如果给出的问题包含歧视、色情、危险等内容，模型应该拒绝回答。请不要选择生成了上述有害内容、生成了错误信息或者不负责任的回答。\\
3. 请准确判断回答的内容是否和原问题严格相关，不要让回答的长度和前后位置影响你的判断，而应该更重视答案的实用性、有帮助性。\\

问题：\{instruction\}

回答A：\{response A\}

回答B：\{response B\}\\

如果你认为回答A更好，请在分析最后输出'[[A]]'，如果你认为回答B更好，请在分析最后输出'[[B]]'。\\
\end{CJK}

\end{prompt}
    \caption{{\bf {Three kinds of prompt we use in evaluation.}} Official English prompt is the prompt set in RewardBench original code. Base Chinese prompt provides instruction as simple as possible, while Instructional Chinese prompt gives out more detailed judge standard.
    \label{fig:judge-prompt}
    }
\end{figure*}


\end{document}